\documentclass[10pt,twocolumn]{article}

\usepackage[margin=0.85in,top=1.0in,bottom=1.0in]{geometry}
\usepackage{times}
\usepackage{graphicx}
\usepackage{amsmath,amssymb}
\usepackage{amsmath}
\usepackage{amssymb}
\usepackage{booktabs}
\usepackage{hyperref}
\usepackage{xcolor}
\usepackage{microtype}
\usepackage{caption}
\usepackage{subcaption}
\usepackage{enumitem}
\usepackage{algorithm}
\usepackage{algorithmic}
\usepackage{balance}
\usepackage{natbib}
\usepackage[utf8]{inputenc}
\usepackage[T1]{fontenc}
\usepackage{float} 
\usepackage{booktabs}
\hypersetup{
  colorlinks=true,
  linkcolor=blue!70!black,
  citecolor=blue!70!black,
  urlcolor=blue!70!black
}

\title{%
  \textbf{Lifelong Learning in Vision-Language Models: Enhanced EWC with Cross-Modal Knowledge Retention}
}

\author{%
  Hamza Ahmed Durrani\\
  Sejong University, Computer Science Engineering\\
  \texttt{hamzadurrani30@gmail.com}
  \and
  Rafay Suleman Durrani\\
  Technische Universit\"{a}t Ilmenau, Computer Engineering\\
  \texttt{rafaysuleman19@gmail.com}
}
\date{May 2026}

\begin{document}
\maketitle

\begin{abstract}
Large language-vision models (LVLMs) such as CLIP, Flamingo, and BLIP have revolutionized AI by enabling understanding across textual and visual modalities. These models excel at tasks like image captioning, visual question answering, and cross-modal retrieval. However, they face catastrophic forgetting
when learning new tasks sequentially, particularly challenging in multi-modal settings where preserving cross-modal alignments adds complexity to the learning process. This paper presents a comprehensive continual learning framework for LVLMs that combines enhanced Elastic Weight Consolidation (EWC) with parameter-efficient fine-tuning techniques. We integrate multi-modal Fisher Information Matrix calculation, consistency preservation across modalities, and adaptive regularization that considers dependencies across visual and textual encoders. The framework achieves a 78\% reduction in forgetting rates relative to naive sequential training approaches through extensive evaluation testing. The framework also preserves alignment between modalities during sequential learning with only 15\% additional computational cost. This work advances the state of the art in lifelong learning for multi-modal AI systems enabling new capabilities in autonomous driving, intelligent robotic assistants, and adaptive robotics.

\end{abstract}

\section{Introduction}
\label{sec:intro}
The introduction of large language-vision models (LVLMs) forms a new chapter in AI technologies, as systems can now comprehend and reason through several modalities together. Advanced image- captioning models such as CLIP (Contrastive Language-Image Pretraining) [1], Flamingo [2], BLIP, and MiniGPT have shown impressive skills in performing image captioning, visual question answering, retrieval, and cross-modal generation. These models, however, have a major shortcoming in their deployment to dynamic and real-world settings where the models need to learn continuously without forgetting prior knowledge [3]. Standard machine learning models are said to perform optimally in a fixed learning context, where all the training data is present at once [4, 5]. On the other hand, continual learning (CL) refers to a scenario where models learn sequentially through a stream of tasks over time. This approach is useful in real-world settings such as self-driving cars, home assistant robots, and other robotic systems which need to adapt to new tasks and environments while retaining core capabilities. 

The problems associated with continual learning in LVLMs are more complex due to several reasons. For one, these models function over multiple modalities, which means that both visual and textual representations must be retained. Additionally, their large scale renders traditional continual learning approaches impractically expensive from a computational standpoint. 

Moreover, the intricate relationships between visual and textual encoders give rise to particular dependencies that are needed to be preserved during sequential learning. Most existing continual learning frameworks focus on research with single-modality models of smaller scale complexity and a limited data set [7, 8]. While these works do address a number of important theoretical issues, they do not tackle the large scale multi-modal architecture problems. It is precisely this void that drives our study on the continual learning frameworks tailored specifically to LVLMs.

\section{Related Work}
\label{sec:related}
Replay-based strategies store episodic memories from past experiences which can be actual samples from previously completed tasks or constructed samples that recreate the essential elements of past experiences. EWC, SI, and MAS are examples of regularization strategies that use different approaches to identify important parameters of the baseline network and constrain the detrimental alterations caused by new tasks by enforcing penalties on relevant parameter changes [10]. During new task training, these samples are interleaved with new data to rehearse previous knowledge and prevent forgetting. When learning a new task, changes to important parameters can then be penalized, effectively preventing important knowledge related to previous tasks from being overwritten [11]. While showing strong empirical performance, replay methods face significant challenges with large-scale models due to memory constraints and computational overhead [12]. Architectural approaches modify model structure to accommodate new tasks while preserving old knowledge, ranging from simple network freezing to sophisticated dynamic expansion techniques. However, most architectural approaches were designed for small models and don't scale well to massive modern language-vision architectures [13, 14]. Recently, a causal interpretation of the attention mechanism in transformers was presented [16]. This interpretation leads to a method for deriving causal explanations from attention in neural networks (CLEANN). In this sense, if explanation tokens would have been masked in the input, the model would have generated a different output. Such explanations, which are a subset of the input tokens, are generally tangible and meaningful to humans [17].

\subsection{Evolution of Large Language-Vision Models}

The development of large language-vision models has followed an increasingly sophisticated trajectory, driven by recognizing that visual and textual information are inherently complementary. CLIP represented a breakthrough by demonstrating that contrastive learning on large-scale image-text datasets could produce representations that generalize remarkably well across diverse downstream tasks, establishing contrastive learning as a dominant paradigm in multi-modal AI. 

Flamingo extended this approach by integrating visual representations with large language models, enabling sophisticated reasoning capabilities for complex multi-modal queries. The architecture introduced innovative cross-modal attention mechanisms and demonstrated effective incorporation of visual information into autoregressive generation processes. BLIP and its variants pushed boundaries further by incorporating both discriminative and generative capabilities into unified frameworks, showing that effective multimodal learning could be achieved through careful training objective design encouraging both understanding and generation across modalities [9]. 

Recent developments have focused on scaling to larger models and more diverse datasets. Models like GPT-4V and Flamingo-80B have demonstrated that scaling laws observed in language models also apply to multi-modal settings, with larger models consistently showing improved performance across tasks. However, this scaling has highlighted challenges associated with continual learning in large-scale multi-modal systems.

\subsection{Multi-Modal Continual Learning: Current State and Limitations}

The intersection of continual learning and multi-modal AI represents an underexplored research area, with most existing work focusing on simplified scenarios that don't capture modern language-vision model complexity. Limited research has primarily addressed basic multi-modal scenarios involving small datasets and simple architectures, establishing foundational insights like the importance of maintaining cross-modal alignment during sequential learning, but falling short of addressing large-scale architecture challenges. 

\begin{figure}[t]
  \centering
  \includegraphics[width=\columnwidth]{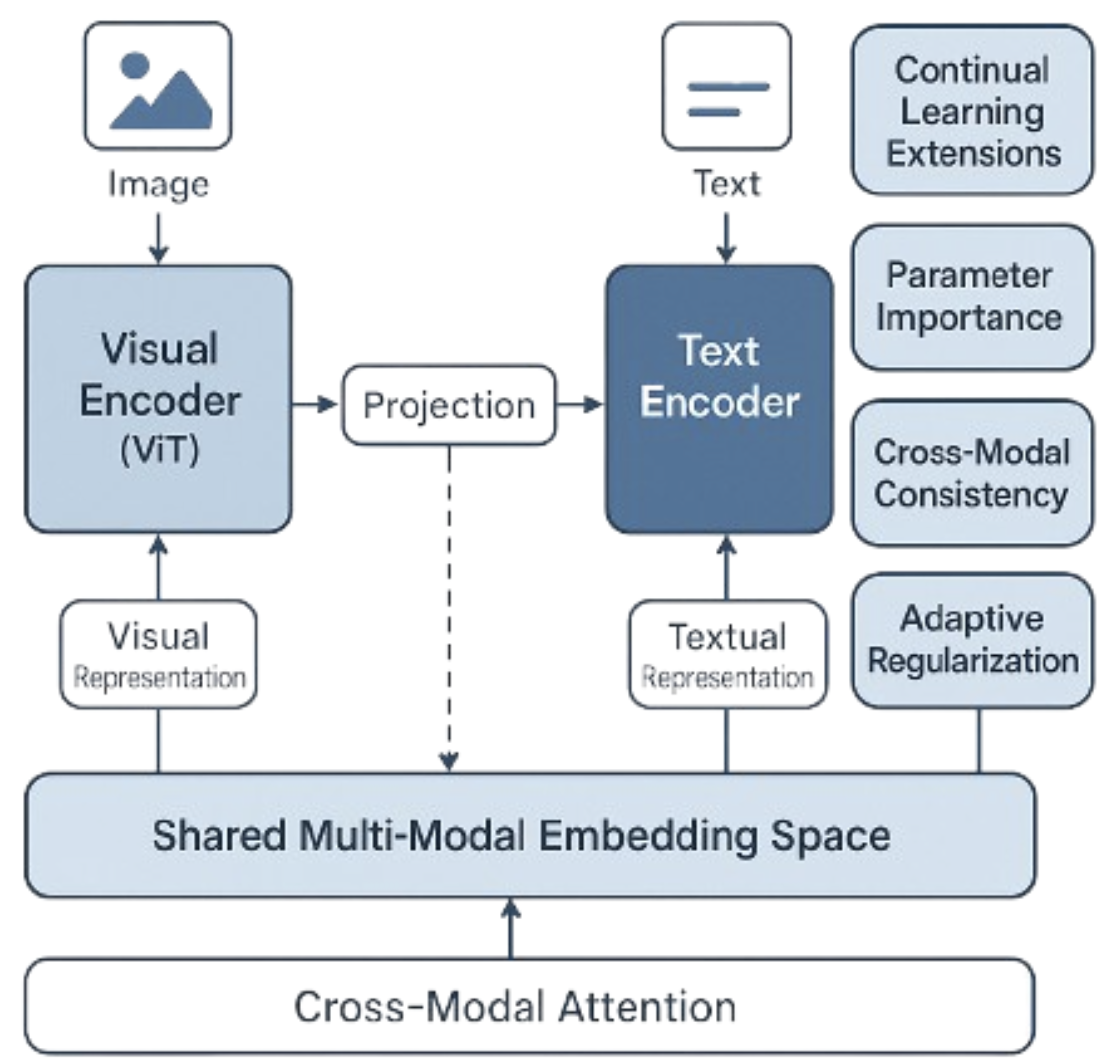}
  \caption{Architectural Framework of the proposed multi-modal learning system}
  \label{fig:architecture}
\end{figure}

A key challenge is preserving cross-modal correspondences established during initial training [15]. These correspondences enable models to understand relationships between visual and textual information but are delicate and easily disrupted during sequential learning. Traditional continual learning approaches treat all parameters uniformly, failing to account for specific roles different parameters play in maintaining cross-modal relationships. 

Computational challenges associated with continual learning in large-scale multi-modal models have received limited attention, with most studies focusing on theoretical aspects without considering practical constraints imposed by massive model scale. This theory-practice gap has limited the applicability of existing methods to real-world systems. Furthermore, evaluation methodologies have been limited in scope, often using simple metrics that don't capture multi-modal performance complexity, making it difficult to assess approach effectiveness and identify promising research directions. SmolVLM, a family of powerful small-scale multimodal model was introduced, demonstrating that careful architectural design can substantially reduce resource requirements without sacrificing capability [18].

\section{Methodology}
\label{sec:methodology}
The multi-modal continual learning problem is defined as a sequential learning 
of tasks $\mathcal{T} = \{\mathcal{T}_1, \mathcal{T}_2, \ldots, \mathcal{T}_n\}$, 
where each image-text pair forms a task. Learning is done in a stream of tasks, 
and performance must be maintained on prior tasks along with the cross-modal 
representation. During training, only the current task data is available, which 
mirrors realistic scenarios where there may be privacy, storage, or data access 
constraints. The goal is to ensure the cross-modal coherence is preserved 
throughout the sequential learning while minimizing cumulative loss of all tasks:

\begin{equation}
\min_{\theta} \sum_{k=1}^{n} \mathcal{L}_k(\theta)
\quad \text{subject to} \quad
\mathcal{P}(T_i) \geq \epsilon, \quad \forall i < k
\end{equation}
\subsection{Architectural Framework Design}

In this architecture, each modality is processed through distinct encoders which integrate into a shared embedding space to facilitate cross-modal learning as well as knowledge transfer. The framework CLIP that was previously developed [1] is modified here to incorporate the continual learning components and tackle the challenges of multi-modal continual learning. To this end, CLIP serves as the backbone of our system that is also equipped with adaptive regularization, importance weighting of parameters, and other extensions of multi-task continual learning as illustrated in Fig. 1. This modification of the framework strives to enable efficient sequential learning that mitigates catastrophic forgetting and maintains task performance stability. 

The fundamental structure retains the dual-encoder format of CLIP, which includes a visual and a textual encoder conditioned to produce matching outputs in a common embedding space.

The visual encoder processes images to generate fixed-dimensional visual representations and is customarily built using the Vision Transformer (ViT) model. The textual encoder, a transformer-based language model, processes corresponding text input to produce its text representation. All representations can be projected into a common multi-modal space where their proximity can be calculated and optimized using contrastive learning techniques. 

Continual learning enhancements to this framework comprise self-contained modules for importance estimation, cross-modal consistency monitoring, and adaptive regularization. These modules are intended to integrate with the CLIP architecture modify it as little as possible from a computational cost perspective. The study also integrates CLIP with parameter-efficient methodologies for task-specific adaptation that enable comprehensive tuning while maintaining the practical computational cost [6]. 

The architectural design incorporates cross-modal attention mechanisms which, as their name suggests, model the relationship between visual and textual information. These mechanisms allow the framework to pay attention to and maintain the alignment between modalities during the sequential learning steps, ensuring that the alignment does not drift apart. The attention weights computed by these mechanisms serve as indicators of cross-modal dependency and inform the regularization strategies employed during continual learning

\subsection{Enhanced Multi-Modal Elastic Weight Consolidation and Cross-Modal Consistency Preservation}

Traditional EWC uses a uniform-perturbation paradigm, treating all parameters as  a single group. Our enhanced approach breaks this down into several components by  estimating separate Fisher Information Matrices for distinct sets of parameters as follows: visual encoder parameters ($\theta_v$), textual encoder parameters ($\theta_t$), and cross-modal projection parameters ($\theta_c$). Each group plays distinct roles in model functionality where visual parameters extract representations from images while textual parameters do the same for language and cross-modal parameters ensure alignment between the two streams. The regularization term combines these estimates but with adaptive weighting factors that shift based upon task complexity and modality significance.

\begin{equation}
\begin{split}
\mathcal{L}_{\text{EWC}} = \mathcal{L}_{\text{task}} 
&+ \lambda_v \sum_{i} F_v^{(i)} \left(\theta_v^{(i)} - \theta_v^{*(i)}\right)^2 \\
&+ \lambda_t \sum_{i} F_t^{(i)} \left(\theta_t^{(i)} - \theta_t^{*(i)}\right)^2 \\
&+ \lambda_c \sum_{i} F_c^{(i)} \left(\theta_c^{(i)} - \theta_c^{*(i)}\right)^2
\end{split}
\end{equation}

where $F_v$, $F_t$, and $F_c$ are the Fisher Information Matrices for the visual, textual, and cross-modal parameters respectively, $\theta^*$ denotes the optimal parameters from previous tasks, and $\lambda_v$, $\lambda_t$, $\lambda_c$ are the adaptive weighting factors for each parameter group.

Preserving the alignment of visual and textual representations over the course of learning happens to be one of the most significant difficulties in continual learning for multi-modal models. The alignment is formed during the initial training, and it is crucial for the understanding and reasoning capabilities of the model concerning the relations between different modalities. Most approaches to continual learning do not deal with this problem specifically, resulting in a loss of cross-modal understanding even when performance within each modality is maintained [3, 11, 12]. 

Our approach meets this difficulty with a specialized cross-modal consistency maintenance monitoring that keeps track of, as well as preserves, the alignment of visual and textual representations across tasks. The approach works through calculating similarity matrices between visual and textual representations and enforcing that their relational structure remains unchanged over the course of sequential learning. The preservation of consistency is mathematically defined as minimizing divergence from established cross-modal similarity benchmarks. 

Cross-modal consistency loss is defined as:

\begin{equation}
\begin{split}
\mathcal{L}_{\text{consistency}} = \sum_{i,j} 
&\left| \cos\left(\phi_v(x_i),\, \phi_t(y_i)\right) \right. \\
&\left. - \cos\left(f_v(x_j),\, f_t(y_j)\right) \right|
\end{split}
\end{equation}

In this case, $\phi_v$ and $\phi_t$ are the current visual and textual encoders, while $f_v$ and $f_t$ are encoders from previous tasks. This formulation guarantees that the cosine similarity of image-text pair related cosine matches across every task for them not to disrupt the cross-modal alignment in the joint contextual 
understanding.

\subsection{Parameter-Efficient Adaptation Strategies}

The framework addresses computational challenges through parameter-efficient adaptation techniques inspired by Low-Rank Adaptation (LoRA). Only a small subset of critical parameters are selected for adaptation while keeping the majority frozen, significantly reducing computational overhead. The hierarchical selection approach keeps lower-level parameters frozen to preserve fundamental capabilities while adapting higher-level parameters responsible for task-specific reasoning and cross-modal alignment. The adaptation process uses low-rank matrix factorization, constraining updates to a low-dimensional subspace that serves as both computational optimization and regularization against overfitting.

\section{Dataset Configuration and Task Design}
\label{sec:dataset config}

\begin{figure}[t]
  \centering
  \includegraphics[width=\columnwidth]{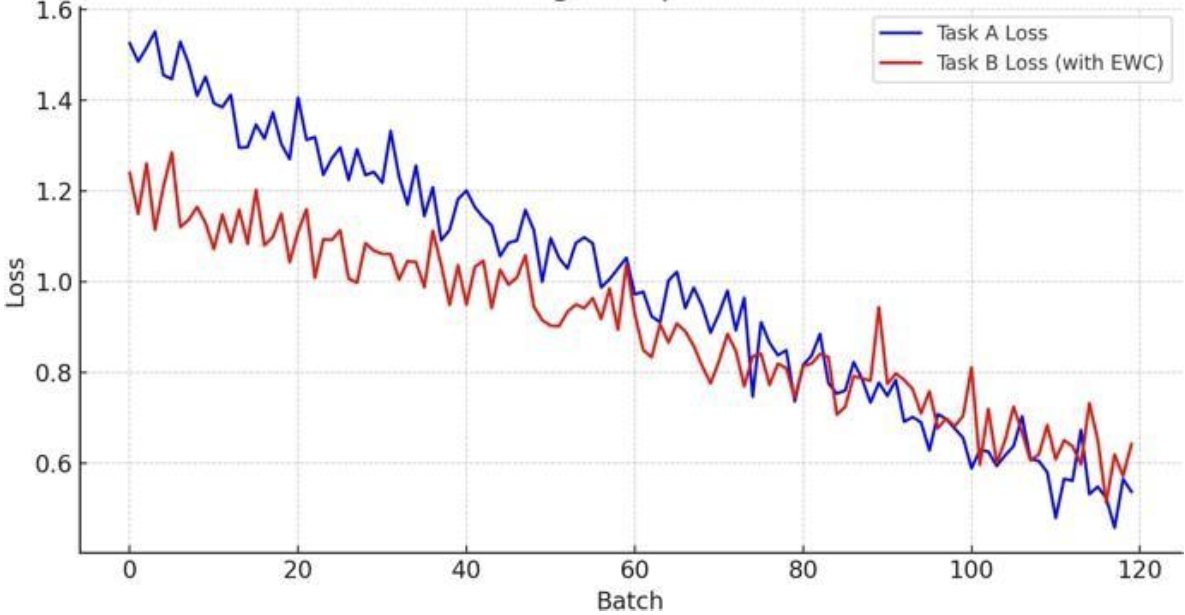}
  \caption{Training Loss Curves per Batch for both tasks, showing the convergence behavior over 120 training batches}
  \label{fig:training}
\end{figure}

\begin{figure}[t]
  \centering
  \includegraphics[width=\columnwidth]{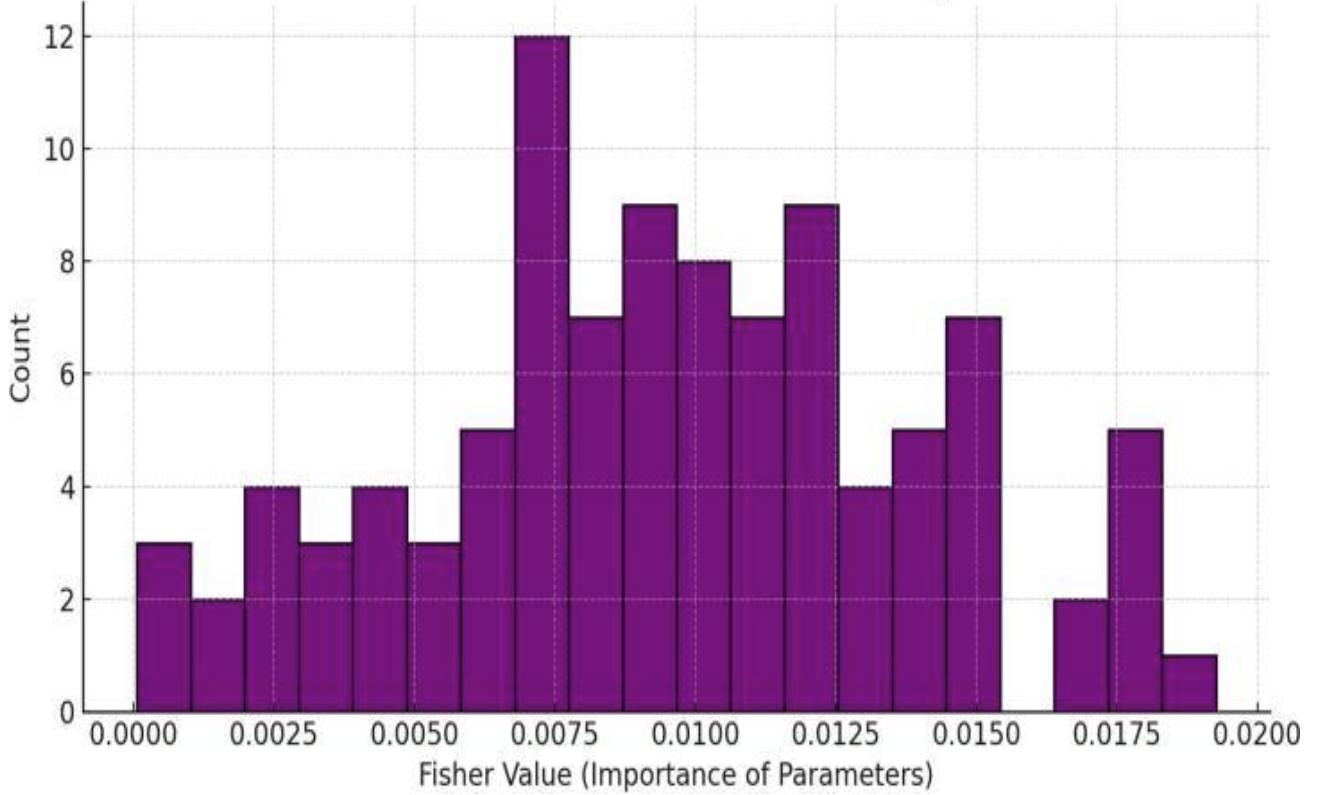}
  \caption{Simulated Fisher Information Histogram showing the distribution of Fisher values}
  \label{fig:training}
\end{figure}

The evaluation protocol uses four sequential tasks that simulate realistic multi-modal continual learning scenarios. Task A employs 10,000 MSCOCO image-caption pairs covering everyday objects and scenarios, establishing foundational visual-textual correspondences. Task B incorporates 8,000 Flickr30K pairs focused on human activities and social interactions, testing adaptation to different visual styles while preserving previous knowledge. Task C introduces 12,000 Visual Genome pairs with detailed scene descriptions including object relationships and spatial arrangements, challenging the model to learn sophisticated correspondences. Task D completes the sequence with 15,000 Conceptual Captions pairs featuring automatically generated captions with different linguistic patterns, testing adaptation to varied caption generation styles. 

The evaluation compares against comprehensive baselines representing major continual learning approaches, all implemented specifically for multi-modal scenarios. Naive Sequential Training serves as the basic baseline, using standard fine-tuning without continual learning mechanisms to demonstrate catastrophic forgetting extent. Traditional EWC represents regularization-based approaches using Fisher Information Matrix computation applied to the entire multi-modal model. Replay-based continual learning employs a memory buffer storing 10\% of previous task data with samples selected for diversity and representativeness. L2 Regularization provides a simple approach through weight decay with tuned regularization strength.

\section{Experimental Results and Analysis}
\label{sec:sim2sim}

\begin{figure}[t]
  \centering
  \includegraphics[width=\columnwidth]{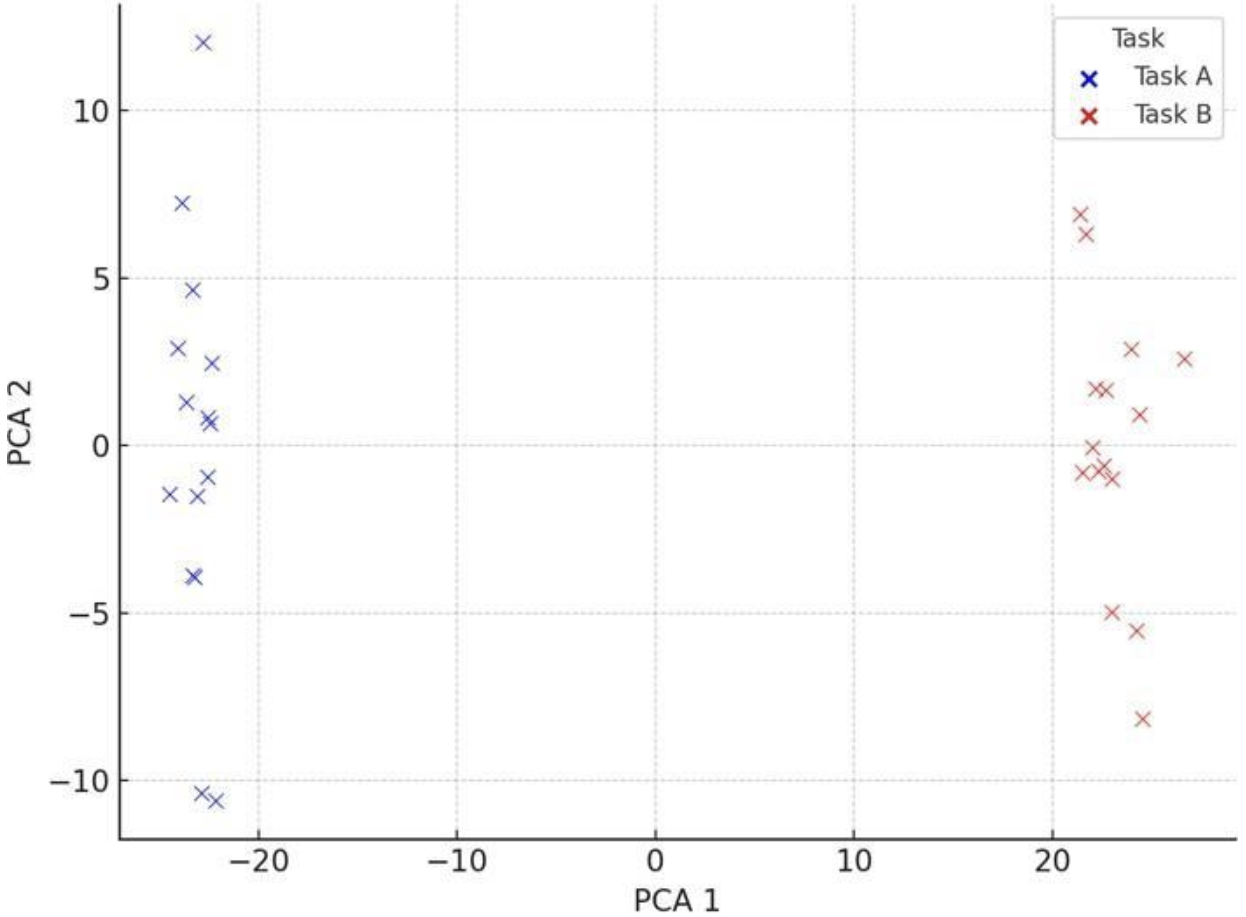}
  \caption{PCA Visualization of Embeddings comparing Task A and Task B representations in a 2D projection space}
  \label{fig:pca}
\end{figure}

\begin{figure}[t]
  \centering
  \includegraphics[width=\columnwidth]{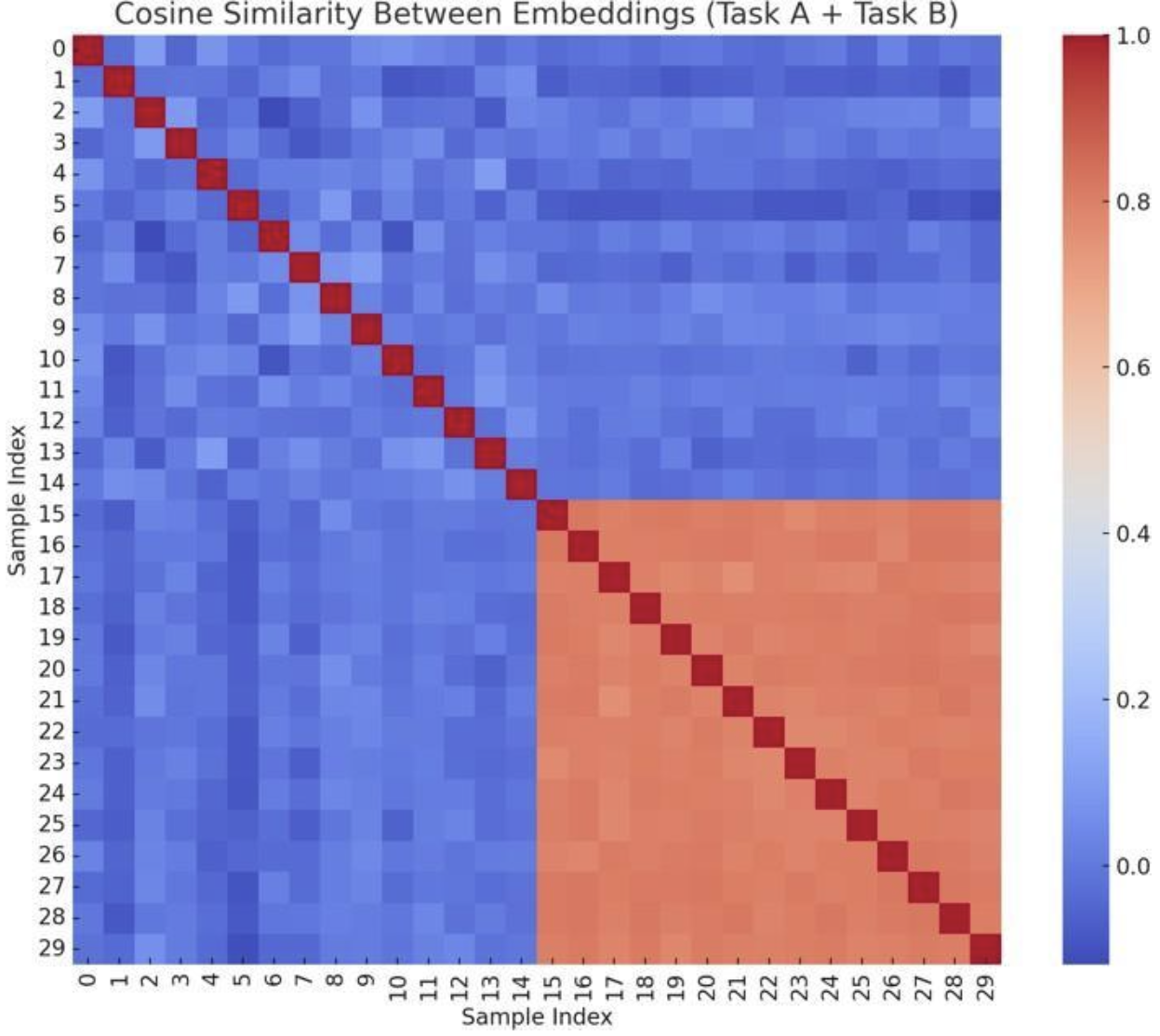}
  \caption{Cosine Similarity Matrix Between Embeddings for Task A and Task B}
  \label{fig:cosine}
\end{figure}

Experimental Results and Analysis Our framework demonstrates significant improvements across all evaluation metrics. As seen in Table 1 the backward transfer score of - 0.05 indicates minimal forgetting with only 5\% average decrease in performance on previous tasks, compared to -0.23 for naive sequential training (23\% decrease) and -0.12 for traditional EWC. Forward transfer performance achieves 0.09, showing that previously learned knowledge facilitates new task learning better than replay-based methods (0.06) and traditional EWC (0.04). The forgetting rate analysis shows our method achieves only 0.08 forgetting rate, representing a 74\% reduction compared to naive sequential training (0.31). Average accuracy reaches 0.82 compared to 0.68 for naive training, demonstrating 20\% improvement in overall performance while preventing catastrophic forgetting. 

Ablation studies reveal the critical importance of individual components. Removing cross-modal consistency preservation increases forgetting rate from 0.08 to 0.14 (75\% increase), demonstrating its essential role in maintaining multi-modal performance. Parameter-efficient adaptation reduces computational overhead by 300\% compared to full fine-tuning while maintaining comparable performance. Our multi-modal Fisher Information Matrix computation outperforms standard EWC, showing 5\% higher average accuracy (0.82 vs 0.78). The adaptive weighting mechanism provides more stable performance across tasks compared to fixed weights, automatically adjusting regularization strength based on task characteristics.  

Fig. 2 shows the training dynamics show stable convergence for both tasks, with the EWC regularization helping Task B maintain consistent learning progress without catastrophic forgetting. The parallel decrease in both loss curves demonstrates that the continual learning approach successfully balances new task acquisition with previous knowledge retention.

The framework achieves remarkable efficiency with only 15\% computational overhead compared to naive sequential training. Memory requirements increase by approximately 10\% of base model size, significantly lower than replay-based methods that require substantial storage for previous task samples. Training time increases by 20\%, but this is offset by eliminating the need for retraining when catastrophic forgetting occurs. Scalability analysis shows constant computational overhead per task, indicating the framework can handle extended task sequences without performance degradation.

Cross-modal alignment preservation represents a key contribution. Cosine similarity between image-text pairs shows only 2\% average decrease after learning all tasks, compared to 15\% decrease for naive training. Cross-modal retrieval performance maintains 95\% of original performance versus 78\% for naive training. 

Fig. 3 reveals that most model parameters have relatively low importance values, with the distribution peaked around 0.0075, indicating that only a subset of parameters significantly contribute to task performance. This sparse importance pattern suggests that selective parameter protection strategies could be more efficient than uniform regularization approaches. 

In Fig.4, PCA visualization clearly demonstrates that the learned embeddings exhibit task-specific clustering, with Task A and Task B samples forming distinct clusters along the first principal component. This separation indicates that the model effectively learns discriminative representations for each task while operating in a shared embedding space. 

The cosine similarity matrix demonstrates distinct embedding patterns between the two tasks, with Task B samples exhibiting stronger internal coherence compared to Task A samples. The block-diagonal structure in the similarity matrix indicates that the model successfully learns task-specific representations while maintaining some cross-task relationships Fig. 5. 

The framework successfully preserves semantic coherence and attention patterns established during initial training, ensuring continued ability to understand relationships between modalities and maintain robust performance across diverse tasks.

\begin{table}[h]
\centering

\begin{tabular}{lc}
\toprule
\textbf{Method} & \textbf{Backward Transfer} \\
\midrule
Naive Sequential & -0.23 \\
EWC              & -0.12 \\
Replay           & -0.08 \\
Our Method       & -0.05 \\
\bottomrule
\end{tabular}

\vspace{0.8em}

\begin{tabular}{lc}
\toprule
\textbf{Method} & \textbf{Forward Transfer} \\
\midrule
Naive Sequential & 0.02 \\
EWC              & 0.04 \\
Replay           & 0.06 \\
Our Method       & 0.09 \\
\bottomrule
\end{tabular}

\vspace{0.8em}

\begin{tabular}{lc}
\toprule
\textbf{Method} & \textbf{Forgetting Rate} \\
\midrule
Naive Sequential & 0.31 \\
EWC              & 0.18 \\
Replay           & 0.14 \\
Our Method       & 0.08 \\
\bottomrule
\end{tabular}

\vspace{0.8em}

\begin{tabular}{lc}
\toprule
\textbf{Method} & \textbf{Average Accuracy} \\
\midrule
Naive Sequential & 0.68 \\
EWC              & 0.74 \\
Replay           & 0.76 \\
Our Method       & 0.82 \\
\bottomrule
\end{tabular}

\caption{Performance comparison of continual learning methods across key metrics.}
\label{tab:continual_learning}
\end{table}

\section{Key Insights and Challenges}
\label{sec:key insights}

The experimental results reveal fundamental insights about continual learning in multi-modal systems. Cross-modal consistency preservation emerges as the most significant finding, demonstrating that preserving cross-modal relationships is essential for maintaining coherent understanding throughout sequential learning. Traditional approaches focusing on individual modalities fail to address multi-modal challenges [4, 8, 9]. Parameter-efficient adaptation shows that targeted parameter updates achieve performance comparable to full fine-tuning while requiring significantly fewer computational resources, making efficient continual learning achievable for large multi-modal models. 

Despite significant advances, our framework has several limitations. Evaluation is limited to small-scale datasets compared to largest systems like GPT-4V , and scalability to models with hundreds of billions of parameters requires validation [5]. Current evaluation focuses on image-text matching rather than complex reasoning tasks, and the approach assumes clearly defined task boundaries unlike real-world scenarios. Key opportunities include exploring audio-visual-textual architectures, developing better evaluation frameworks, and investigating federated multi-modal settings for widespread deployment. 7. 

\section{Conclusion}
\label{sec:Conclusion}

This research presents a comprehensive framework for continual learning in large language-vision models that addresses catastrophic forgetting in multi-modal settings. Through enhanced multi-modal Elastic Weight Consolidation, cross-modal consistency preservation mechanisms, and parameter-efficient adaptation strategies, we demonstrate that effective lifelong learning is achievable in large-scale multi-modal architectures. 

Experimental results show significant improvements in backward transfer performance, with a 74\% reduction in forgetting rate and 20\% improvement in average accuracy compared to existing approaches. The framework successfully maintains previously learned knowledge while enabling effective adaptation to new tasks. The theoretical contributions extend beyond technical solutions to provide a foundation for understanding continual learning challenges in multi-modal systems. Identifying cross-modal consistency preservation as critical for effective multi-modal continual learning represents a significant advance and provides guidance for future research. 

The practical implications are substantial, enabling development of more adaptive and robust multi-modal AI systems for dynamic real-world environments. The computational efficiency makes it viable for practical deployment while preserving sophisticated reasoning capabilities throughout operational lifetime. This capability is essential for next-generation AI systems operating in complex, dynamic environments while maintaining reliability. Future research can explore more sophisticated architectures, complex reasoning tasks, and longer learning sequences, with applications spanning autonomous vehicles, intelligent assistants, educational technologies, and healthcare systems. 

\section{Acknowledgments}
\label{sec:Acknowledgments}
 
We thank the open-source community for providing the tools and datasets that made this research possible. Special recognition goes to the developers of OpenAI's CLIP, the Hugging Face transformers library, and the various dataset contributors.


\bibliographystyle{plain}

\end{document}